\documentclass[10pt,twocolumn,letterpaper]{article}

\usepackage{cvpr}              
\usepackage{soul}
\usepackage{float}
\usepackage{cuted}
\usepackage{multirow}
\usepackage{tabularx}
\definecolor{cvprblue}{rgb}{0.21,0.49,0.74}
\usepackage[pagebackref,breaklinks,colorlinks,allcolors=cvprblue]{hyperref}

\def\paperID{} 
\def\confName{CVPR}
\def\confYear{2026}

\title{Making Avatars Interact: Towards Text-Driven Human-Object Interaction for Controllable Talking Avatars}

\author{Youliang Zhang$^{1,2}$ \textsuperscript{\textdagger}
    \hspace{0.15in}
    Zhengguang Zhou$^{2}$ \textsuperscript{\textdagger}
    \hspace{0.15in}
    Zhentao Yu$^{2}$ 
    \hspace{0.15in}
    Ziyao Huang $^{2}$ \\
    \hspace{0.15in}
    Teng Hu$^{2}$
    \hspace{0.15in}
    Sen Liang$^{2}$
    \hspace{0.15in}
    Guozhen Zhang$^{2}$
    \hspace{0.15in}
    Ziqiao Peng$^{2}$
    \hspace{0.15in}
    Shunkai Li$^{2}$ \\
    \hspace{0.15in}
    Yi Chen$^{2}$
    \hspace{0.15in}
    Zixiang Zhou$^{2}$
    \hspace{0.15in}
    Yuan Zhou$^{2}$ \textsuperscript{\textsection}
    \hspace{0.15in}
    Qinglin Lu$^{2}$ \textsuperscript{\textdaggerdbl}
    \hspace{0.15in}
    Xiu Li$^{1}$\textsuperscript{\textdaggerdbl}\\
    \\
    $^1$Tsinghua University.\hspace{0.15in} $^2$Tencent HY.\\
    }

\begin{document}

\maketitle


\begingroup
\renewcommand\thefootnote{}
\footnotetext{
    \textdagger~Equal Contribution \quad 
    \textdaggerdbl~Corresponding author \quad 
    \textsection~Project Leader
}
\footnotetext{
    *~Work done during an internship at Tencent HY.
}
\endgroup

\begin{strip}
\centering
    \vspace{-50pt}
    \includegraphics[width=\textwidth]{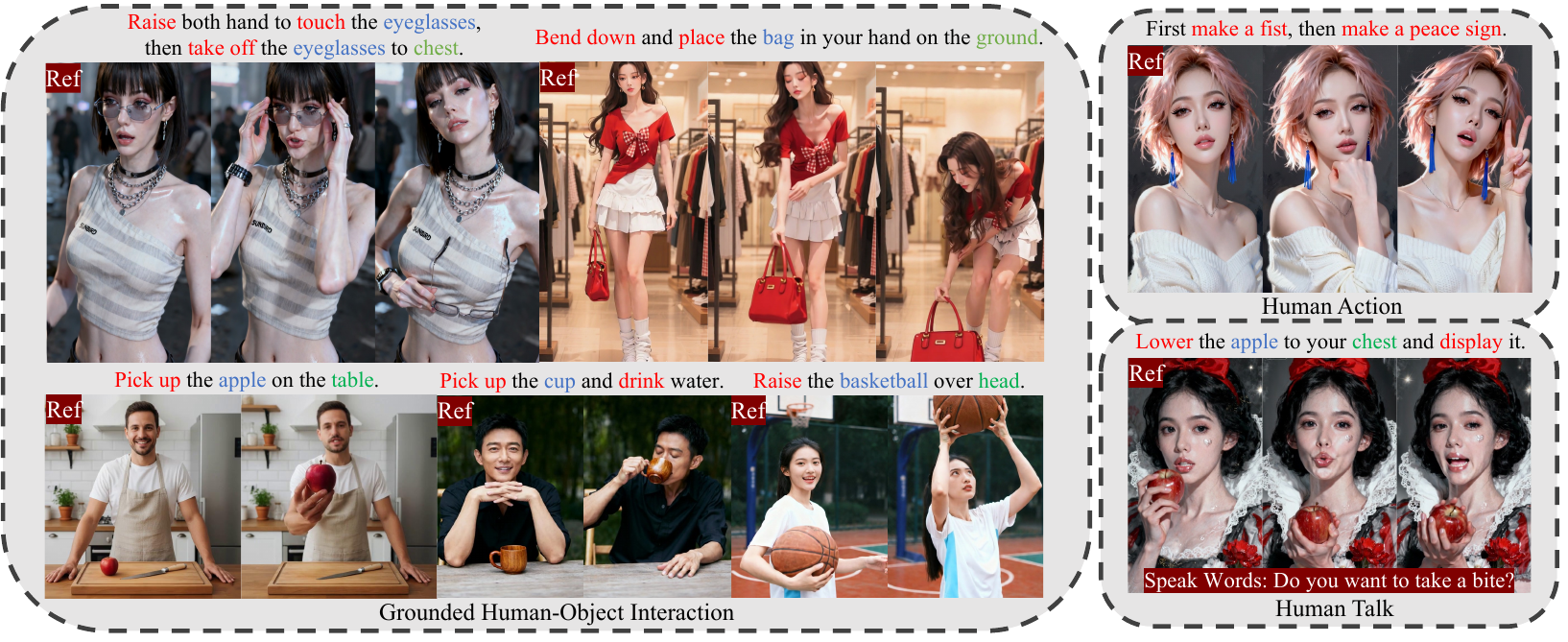}
    \vspace{-22pt}
    \captionof{figure}{\textbf{Input text, audio, and reference image, our method can generate human videos that can talk, act, and interact with objects. Grounded means that all action and object interaction occurs within the environment provided by the reference image. } }
    \label{fig:motivation}
\end{strip}

\vspace{-35pt}
\begin{abstract}

Generating talking avatars is a fundamental task in video generation. Although existing methods can generate full-body talking avatars with simple human motion, extending this task to grounded human-object interaction (GHOI) remains an open challenge, requiring the avatar to perform text-aligned interactions with surrounding objects. This challenge stems from the need for environmental perception and the control-quality dilemma in GHOI generation.
To address this, we propose a novel dual-stream framework, \textbf{InteractAvatar}, which decouples perception and planning from video synthesis for grounded human-object interaction. Leveraging detection to enhance environmental perception, we introduce a Perception and Interaction Module (PIM) to generate text-aligned interaction motions. Additionally, an Audio-Interaction Aware Generation Module (AIM) is proposed to synthesize vivid talking avatars performing object interactions.
With a specially designed motion-to-video aligner, PIM and AIM share a similar network structure and enable parallel co-generation of motions and plausible videos, effectively mitigating the control-quality dilemma. Finally, we establish a benchmark, \textbf{GroundedInter}, for evaluating GHOI video generation. Extensive experiments and comparisons demonstrate the effectiveness of our method in generating grounded human-object interactions for talking avatars.
Project page: \href{https://interactavatar.github.io/}{https://interactavatar.github.io/}.
\end{abstract}    
\vspace{-3mm}
\section{Introduction}
\label{sec:intro}

Human-centric video generation has achieved remarkable progress, with significant strides made in tasks like photorealistic talking head synthesis, focusing primarily on accurate lip-synchronization and facial expression generation~\cite{wang2024v,xu2024hallo,meng2025echomimicv2,chen2025hunyuanvideo,tian2024emo}. 
However, to construct vivid and practical digital humans, we must move beyond facial modeling and endow them with the ability to perform complex actions and interact meaningfully with their surroundings.

We introduce \textbf{Grounded Human-Object Interaction} for talking avatars, 
aiming to enhance current digital humans with environmental perception and text-driven human-object interaction generation ability. Different from the current talking avatar and HOI generation task, GHOI features: 1) Environment-aware, perceives its environment and performs reasonable actions within it. 2) Consistent with the initial frame, rather than generating a new scene. 3) Text-following, do not demand explicit pose or object trajectories for control, and 4) Free of additional object conditions, operating objects in the scene based on text.

Despite recent advancements, existing paradigms still fall short of enabling such nuanced, scene-aware interaction due to fundamental architectural limitations:
\textbf{\textit{(1)}} 
\textbf{Audio-driven methods}, through several pioneering works on simple full-body animation~\cite{cui2025hallo3,wang2025fantasytalking,chen2025hunyuanvideo,lin2025omnihuman}, typically learn a direct mapping from acoustic features to pixel space, lacking explicit modeling of objects and environments, thus making complex human-object interactions difficult to control;
\textbf{\textit{(2)}} \textbf{Pose-driven approaches}~\cite{wang2025unianimate,cheng2025wan,tu2025stableanimator,tan2024animate} offer explicit control but offload the planning burden to the user. These methods require pre-defined skeletal sequences as input, which are not only difficult and expensive to obtain but also often misaligned with the specific context of a reference image;
\textbf{\textit{(3)}} \textbf{Subject-consistent methods}~\cite{hu2025hunyuancustom,chen2025humo,liu2025phantom} excel in preserving subject identity and achieving coherent integration, but lack mechanisms for grounded interaction. They are designed to synthesize new videos from text rather than to carry out contextual commands in an interactive environment.
Therefore, enabling a speaking avatar to perform stable, text-driven object interactions remains a significant and unresolved challenge. The core difficulties are twofold.



\textbf{\textit{(1)}} \textbf{Scene-Action Grounding}: 
Instead of merely generating a video featuring human-object interaction, GHOI demands the interaction to occur within a specific environment involving designated objects. This requires the model to interpret the spatial layout and semantic content, and to ground textual commands within this visual context.
\textbf{\textit{(2)}}  \textbf{Control-Quality Dilemma}: 
Owing to the complexity of GHOI, existing methods often face a trade-off between controllability and visual quality. They either generate high-fidelity videos that struggle to follow instructions or achieve reasonable responses at the expense of video fidelity.

To address these challenges, we propose InteractAvatar, a novel dual-stream Diffusion Transformer (DiT) framework designed to generate grounded human-object interactions in talking avatars. This framework explicitly decouples perception and planning from video synthesis.
For clarity, we define \textbf{motion} as the rendered skeletal poses and object boxes.
We introduce a Perception and Interaction Module (PIM) to tackle the grounding challenge, which generates text-aligned motion sequences based on the perception of the reference image.
The PIM is jointly trained on detection and motion generation tasks, ensuring the production of scene-aware and text-aligned motions.
Conditioned on both audio and the motion latent features, an Audio-Interaction-aware Generation Module (AIM) is proposed to synthesize the final video.
PIM and AIM share a similar network architecture and achieve the parallel co-generation of motions and videos.
With a specially designed motion-to-video (M2V) aligner, our symmetrical yet decoupled framework effectively alleviates the control-quality dilemma in complex GHOI video generation, achieving vivid and plausible grounded human-object interactions for talking avatars.

To rigorously evaluate our approach, particularly its ability to model grounded human-object interactions, we constructed the GroundedInter benchmark. GroundedInter comprises 600 distinct test cases, each containing a reference image, a structured textual description of the interaction, and the corresponding speech audio. Extensive experiments conducted on this benchmark demonstrate that our proposed method significantly outperforms existing approaches in generating plausible, controllable, and high-quality grounded human-object interaction videos.

Our main contributions are summarized as follows:
\begin{itemize}
\item We propose \textbf{InteractAvatar}, a novel dual-stream DiT framework that explicitly decouples perception planning from video synthesis, achieving text-driven grounded human-object interaction generation for talking avatars.
\item We introduce a \textbf{Perception and Interaction Module} (PIM), trained on both detection and motion generation tasks, to generate scene-aware and text-aligned motions. In addition, we propose an \textbf{Audio-Interaction Aware Generation Module} (AIM) with an M2V aligner to produce videos with controllability and high visual quality.
\item Beyond the core GHOI generation, our framework provides \textbf{unified and flexible multimodal control}, accepting any combination of text, audio, and motion as input.
\item We establish \textbf{GroundedInter}, a benchmark for fine-grained evaluation of GHOI. Experiments demonstrate that our method significantly outperforms existing approaches in producing controllable high-quality talking avatars with grounded human-object interactions.
\end{itemize}

\section{Related Work} 
\label{sec:related}

\subsection{Audio-driven human animation.}
Given audio and human reference images as input, audio-driven human animation aims to generate videos with well-aligned lip movements. 
Initial research predominantly focused on head-and-shoulder animations, SadTalker~\cite{zhang2023sadtalker} employing 3D motion coefficients to improve realism, Hallo~\cite{xu2024hallo} using a hierarchical diffusion approach for lip-sync accuracy, EchoMimic~\cite{chen2025echomimic} combining audio with facial landmarks to enhance stability, and Loopy~\cite{jiang2024loopy} leveraging temporal modules to generate natural motion without manual templates. Subsequently, research has progressively extended this capability toward half-body and full-body digital human synthesis with body action modeling. Hallo3~\cite{cui2025hallo3} first applied a pre-trained DiT model for portrait animation, 
Huanyuanvideo-Avatar~\cite{chen2025hunyuanvideo} achieves dynamic, emotion-controllable, and multi-character dialogue video generation.
FantasyTalking~\cite{wang2025fantasytalking} proposes a dual-level optimization for global motion and lip synchronization, and OmniHuman-1~\cite{lin2025omnihuman} scales up training data with hybrid motion conditions to generate more realistic full-body videos. However, these methods lack the capacity to generate more complex grounded human-object interactions. 
Our work extends audio-driven animation to incorporate human-object interaction, thereby enhancing the realism and practical applicability of the generated avatars.

\subsection{Human-Object Interaction Video Generation}
Recently, Human-Object Interaction (HOI) video generation has garnered increasing attention. Some works focus on editing existing footage. MIMO~\cite{men2025mimo} and AnimateAnyone2~\cite{hu2025animate} substitute humans within dynamic scenes, whereas HOI-Swap~\cite{xue2024hoi} and ReHoLD~\cite{fan2025re} replace handheld objects. However, a primary limitation of these methods is their inability to animate a static reference image from scratch.
Another works seek fine-grained control via complex conditional inputs. AnchorCrafter~\cite{xu2024anchorcrafter} utilizes pose and depth maps for interaction-aware human motion, ManiVideo~\cite{pang2025manivideo} employs precise 3D models of both the human and object for accurate control, and HunyuanVideo-HOMA~\cite{huang2025hunyuanvideo} explores weak HOI conditions with a novel appearance injection method. Nevertheless, acquiring and aligning these complex conditions with an arbitrary ref image is often impractical, hindering their use in digital human applications. In contrast, our method overcomes these challenges by generating realistic human-object interaction videos conditioned solely on text, thereby eliminating the need for source videos or other complex control signals.

Recently, subject-consistent video generation also exhibited the capacity to synthesize HOI video. Phantom~\cite{liu2025phantom} leverages the inherent consistency of DiT-based models by concatenating reference image latents with video latents. Concurrent works such as InterActHuman~\cite{wang2025interacthuman} and HunyuanCustom~\cite{hu2025hunyuancustom} integrate subject-consistent video generation with audio-driven animation, while HuMo~\cite{chen2025humo} proposes a progressive multimodal training paradigm to enable flexible control. However, a critical distinction between GHOI and subject-consistent video generation is that these methods do not preserve the world defined by the environment of the ref image. Instead, they typically generate an entirely new scene based on text and subject appearance, rather than situating the interaction within the provided environment.

\section{Method}
\label{sec:method}

\begin{figure*}[t]
\vspace{-6mm}
    \centering
\includegraphics[width=\textwidth]{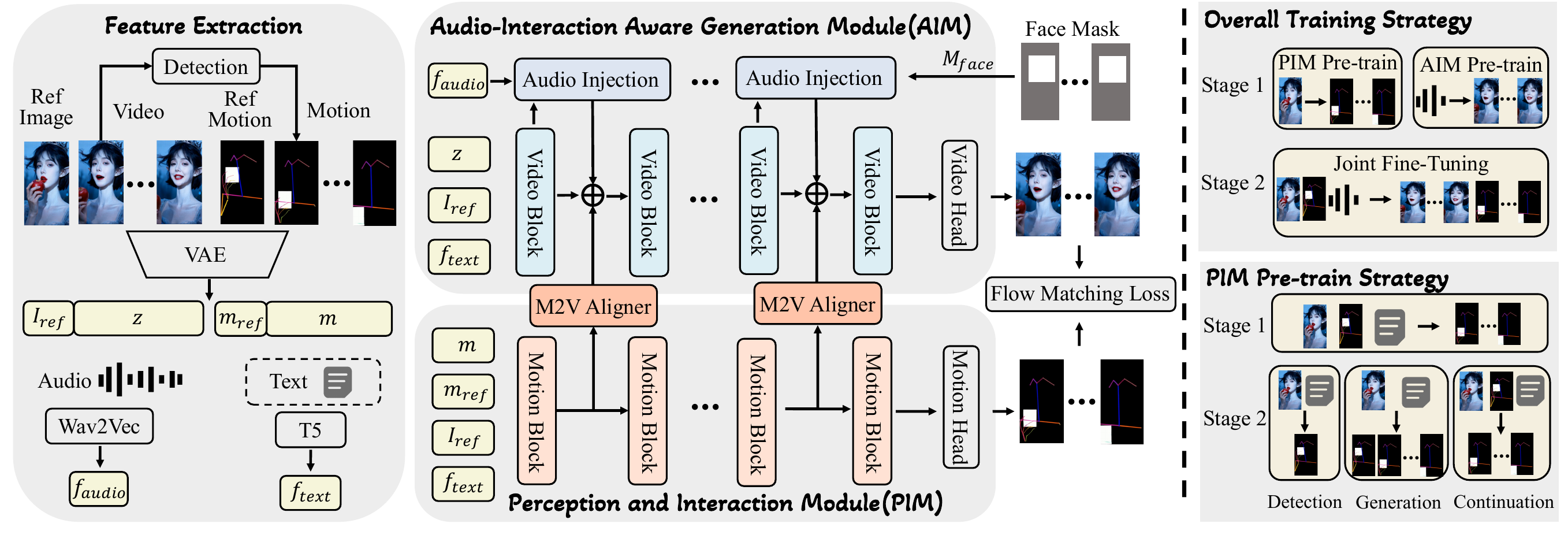}
\vspace{-9mm}
\caption{\label{fig:pipline}\textbf{Overview of our dual-stream InteractAvatar framework and multimodal conditioned training strategy.} 
    }
\vspace{-6mm}
\end{figure*}

As shown in Fig.~\ref{fig:pipline}, we introduce a dual-stream Diffusion Transformer (DiT) for generating text-controlled grounded human-object interactions. Our architecture decomposes this complex task into two sub-problems, perceptual planning and video rendering, handled by the Perception and Interaction Module (PIM) and the Audio-Interaction aware Generation Module (AIM), respectively.
We define \textbf{motion} as the rendered human skeletal poses and object boxes.
A key aspect of our design is the parallel co-generation of structural motion sequences and corresponding video frames. 
The PIM acts as the \textbf{planning brain}, focusing on high-level structure by generating scene-aware and text-aligned motions. The AIM serves as the \textbf{rendering engine}, synthesizing high-fidelity videos with precise audio lip-sync.
Rather than a sequential pipeline, AIM is informed throughout the generation process via the M2V aligner, which employs a layer-wise residual injection mechanism to ensure that the generated video consistently adheres to the evolving structural motion from PIM in lockstep.

Both modules are trained under a unified Flow Matching paradigm. A pre-trained Variational Autoencoder (VAE)~\cite{wan2025} encodes all visual inputs (videos, motions, ref images) into latent space, while a T5~\cite{2020t5} encoder embeds the textual commands. The model \(u_\theta\) is optimized to predict a vector field \(v_t\) with the following objective function:
\begin{equation}
    \mathcal{L}_{\text{FM}} = \mathbb{E}_{t, z_0, c} \left[ \left\| v_t(z_t) - u_\theta(z_t, t, c) \right\|_2^2 \right],
    \label{eq:flow_matching}
\end{equation}
where \(z_0\) is the clean latent from the VAE, \(z_t\) is its noised version at timestep \(t\), \(v_t(z_t)\) is the ground-truth vector field, and \(c\) represents all conditioning information.

\subsection{Perception and Interaction Module}
\label{sec:PIM}

The PIM is responsible for parsing the environmental context from a reference image \(I_{ref}\) and generating a spatio-temporally reasonable interaction motion with the guidance of the text prompt \(T\). We define the interaction motion \(m\) as a composite visual representation, comprising a sequence of human skeletal keypoints, \(P = \{{p}_t\}_{t=0}^{N-1}\), and object bounding box trajectories, \(O = \{{o}_t\}_{t=0}^{N-1}\), where \(N\) is the total number of frames. This motion serves as an intermediate representation that is injected at the feature level into the video generation module AIM, thereby enabling disentangled control over the human-object interaction dynamics.

\textbf{Interaction Motion Generation.}
To precisely control the generation process, we explicitly partition the motion generation into two distinct task types: pure action generation, involving only the human pose sequence \(P\), and HOI generation, including both the pose sequence \(P\) and object trajectories \(O\). 
We inject the task-specific information by concatenating a task embedding vector \({f}_{{task}}\) corresponding to the task type \(c_{task} \in \{\text{ACTION}, \text{HOI}\}\) with the text prompt embedding \({f}_{{text}}\). This creates a unified conditioning vector for the cross-attention layers of the model:
\begin{equation}
{f}_{{text}} = \text{Concat}({f}_{{text}}, {f}_{{task}}).
\end{equation}

The reference image \(I_{ref}\) is temporally prepended to the first motion frame \(m_{ref}\), the model takes \(m_{ref}\) as first frame to generates overall motion sequence \(\{m_t\}_{t=1}^{N-1}\) conditioned on \(I_{ref}\). To ensure the reference image provides effective global guidance without conflicting with the positional encodings of the motion sequence, we employ a custom mapping for its Rotary Position Embeddings (RoPE). For a given patch \(z_I\) from the reference image with an original 3D position index \((l, i, j)\), its position is remapped as:
\begin{equation}
    \text{RoPE}_{z_I}(l, i, j) = \text{RoPE}(-1, i + w, j + h).
\end{equation}
This strategy positions the reference image as a special environment frame at a virtual timestep of -1, allowing it to condition the entire sequence generation without disrupting the temporal integrity of the video frames themselves.
Given that motions encode structure rather than fine-grained texture, we therefore constrain the shorter side of the motions to 256 pixels, preserving essential structural information while significantly reducing computational overhead.



\textbf{Environment Perception Training.}
Naive conditioning on the reference image is insufficient for the model to understand the complex scene layouts and object geometries. To address this, we develop an environment perception training curriculum that encourages the model to engage more deeply with the reference image.
Our training curriculum dynamically alternates between two task modes:
\textbf{(1) Conditional Continuation.} In this mode, the model receives the reference image \(I_{ref}\), the first frame motion \(m_{ref}\), and the text prompt \(T\) as conditions. Its objective is to generate the subsequent motion sequence \(\{m_t\}_{t=1}^{N-1}\). 
\textbf{(2) Perception as Generation.} In this mode, the first frame motion \(m_{ref}\) is masked out from the input. The model is tasked with generating the entire sequence, including the first frame \(m_{ref}\), conditioned solely on \(I_{ref}\) and \(T\). To further intensify the focus on scene perception, we probabilistically set the target video length to 1. This reduces the generation task to a detection-like objective, the model performs a single-frame generation where the ground truth is the motion \(m_{ref}\).

This detection-like task is also optimized using the same flow matching objective, with the loss computation simply confined to the first frame. This unified framework seamlessly trains the PIM to leverage a single set of network parameters for two seemingly disparate yet fundamentally related tasks: parsing scene structure and generating temporal motions from perception. This strategy obviates the need for heterogeneous loss functions and their associated hyperparameters, which may introduce training instabilities. 

\subsection{Audio-Interaction Aware Video Generation}

\textbf{Audio Injection.}
Precise lip synchronization with the input audio is paramount for generating realistic avatars. We employ a pre-trained Wav2Vec model~\cite{baevski2020wav2vec} as our audio feature extractor. To capture the co-articulation and temporal dynamics of speech, rather than relying on isolated phonemes, we extract features from a contextual window around each timestep. For the \(i\)-th video frame, its corresponding audio feature \({f}_{\text{audio}, i}\) is computed by encoding and aggregation:
\begin{equation}
    {f}_{\text{audio}, i} = \text{Aggregate}\left( \text{Wav2Vec}(\mathbf{a}_{[c_i - w : c_i + w]}) \right),
\end{equation}
where \(\mathbf{a}\) is the raw audio, \(c_i\) is the center sample aligned with frame \(i\), \(w\) denotes the half-width of the contextual window, and \(\text{Aggregate}(\cdot)\) is a feature aggregation operation, such as concatenation, for temporal video audio align.

To achieve frame-level audio-visual alignment, the extracted audio feature sequence \(\{{f}_{\text{audio}, i}\}\) is injected into the AIM via cross-attention. To accelerate convergence and explicitly guide this conditioning, especially in early training, we introduce a spatial face mask, \(\mathbf{M}_{\text{face}}\). This mask spatially weights the output of the audio injection module, concentrating its influence on the facial region for stable training:
\begin{equation}
    \mathbf{h}'_{v, l} = \mathbf{h}_{v, l} + \text{CrossAttention}(\mathbf{h}_{v, l}, {f}_{\text{audio}}) \odot \mathbf{M}_{\text{face}},
\end{equation}
where \(\mathbf{h}_{v, l}\) represents the visual features at layer \(l\) of the AIM and \(\odot\) denotes element-wise multiplication. This strategy effectively channels the impact of the audio signal to the target facial area, leading to a more stable learning process.

\textbf{Motion to Video Aligner.}
Leveraging the isomorphic architecture of the PIM and AIM, we introduce the M2V aligner with a layer-wise residual injection mechanism for motion control.
Specifically, we inject the feature residual, the difference between the output of each DiT block in the PIM, into the corresponding layer of the AIM. For given layer \(l > 0\) in the PIM, its residual \(\Delta \mathbf{h}_{m, l}\) is computed as:
\begin{equation}
    \Delta \mathbf{h}_{m, l} = \mathbf{h}_{m, l}^{\text{out}} - \mathbf{h}_{m, l-1}^{\text{out}}.
\end{equation}
For the base layer (\(l=0\)), we use \(\mathbf{h}_{m, 0}^{\text{out}}\) directly.

To reconcile the discrepancy of the low-resolution motion and the high-resolution video, we first upsample the residual feature \(\Delta \mathbf{h}_{m, l}\) to match the spatial dimensions of video features \(\mathbf{h}_{v, l}\) using simple bilinear interpolation. This is followed by a projection through a zero-initialized linear layer. The final output of an AIM block is computed as:
\begin{equation}
    \mathbf{h}_{v, l}^{\text{out}} = \text{Block}_v(\mathbf{h}_{v, l}^{\text{in}} + \text{Linear}_{\text{zero}}(\text{Interp}(\Delta \mathbf{h}_{m, l}))).
\end{equation}

This residual injection strategy significantly reduces loss fluctuations during training. Furthermore, the zero-initialization of the linear layer ensures stability in the early stages of training by allowing the model to gradually learn the guidance of the motion. This effectively prevents visual artifacts, such as the ghosting of the motion's skeletal lines, from leaking into the final generated video.

\textbf{Unified Training for Generative and Driven Control.}
A truly versatile digital human system should not only generate motions from textual prompts but also faithfully follow explicit driving signals, such as skeletal sequences, interaction motions, or object trajectories. To achieve this, we co-train the model on two complementary tasks: Joint Generation and External Driving.
For External Driving, the model is provided with a clean driving signal, such as a skeleton sequence, which is fed directly to the PIM. Here, the PIM actually acts as a feature encoder, which is elegantly achieved by setting the diffusion timestep \(t\) to 0. 
Driven signals are also extracted as layer-wise feature residuals \(\{\Delta \mathbf{h}_{m, l}\}\), which are then injected into the AIM to guide the generation process.
To balance these dual capabilities, we employ a 4:1 training data ratio between the Joint Generation and External Driving tasks. 
This co-training strategy endows our model to function both as a creative, text-guided motion video generator and as a precise, controllable animator, all within a single, unified framework.

\subsection{Multimodal Conditioned Training Strategy}
\label{sec:training_strategy}

We designed a three-stage training curriculum to progressively integrate multimodal conditions, ensuring stable learning and preventing conflicts between modalities.

\textbf{PIM Pre-training for Scene-Aware Planning.}
The initial stage is dedicated to independently training the PIM. The goal is to establish its core ability to parse static scenes from a reference image and generate coherent motion plans based on text. The training curriculum blends multiple tasks: (1) a static detection task to build scene perception, (2) a continuation task, motion continuation from a given first motion $m_{ref}$, and (3) a crucial hybrid generation task that requires generating the motion sequence from only the image $I_{ref}$ and text. We found that this mixed-task approach, which forces a synergy between perception and generation, is superior to the pure generation pipeline. The training strategy of those tasks is shown in Fig. \ref{fig:pipline}.

\textbf{AIM Pre-training for Audio-Visual Synchronization.}
The second stage imbues AIM with the ability for audio-driven lip synchronization while preserving its underlying image-to-video (I2V) quality. Crucially, we found that the training order is paramount. The model should be trained with audio before being exposed to interaction motion. We attribute this to the distinct nature of condition signals: audio is a soft, localized, and heterogeneous modality injected through cross-attention, whereas motion serves as a hard, global, and homogeneous structural constraint introduced via residual injection.
Introducing the dominant structural motion condition only after the weaker local audio conditioning has stabilized effectively prevents the audio signal from being suppressed during training optimization.

\textbf{End-to-End Joint Fine-tuning.}
In the final stage, the pre-trained PIM and AIM are jointly fine-tuned, where the PIM’s outputs are injected into the AIM as layer-wise residuals.
The training strategy adopts a probabilistic mix of tasks and conditions to unify the model’s capability to accommodate any combination of text, audio, and motion inputs.
Specifically, 30\% of the samples are conditioned on audio to preserve lip sync ability, 15\% on ground-truth motions for driven ability, and 60\% on joint video-motion generation.
Retaining a considerable portion of pure image-to-video samples serves as a critical regularizer, improving identity preservation, video dynamics, and generalization.

\section{Experimental Results and Analysis}
\label{sec:expriment}

\subsection{Experimental Details}
Both our PIM and AIM are initialized from the wan2.2-5B~\cite{wan2025}. The PIM is trained for 30,000 steps on 3-10 second clips of skeleton sequences and object trajectories (short side 256px), while the AIM is trained for 5,000 steps on 3-6 second video clips (short side 704px). 
In the initial pre-training phases for PIM and AIM, the learning rate is fixed at 1e-5. This is followed by a joint training phase of 4,000 steps where the learning rate is lowered to 2e-6. 

\subsection{Benchmark Construction}
To facilitate a rigorous evaluation for human object interaction generation, we constructed a benchmark with images generated by jimeng4.0, named GroundInter.
GroundInter benchmark includes 400 images depicting actual or potential human-object interactions with 1-3 objects, with 100 different common objects.
Each image is richly annotated with object types, 1-3 corresponding action descriptions (detailing interaction type, objects, and temporal segments), and matching dialogue scripts synthesized into audio using CosyVoice~\cite{du2024cosyvoice}, resulting in 600 test cases. Additionally, we provide ground-truth object segmentation masks~\cite{kirillov2023segany}, detection results~\cite{zhang2022dino}, and DW-Pose keypoints~\cite{yang2023effective}. We introduce a suite of metrics focused on interaction quality: (1) VLM-QA, we design 30 questions structured around three categories (object, human, interaction), and the VLM~\cite{wang2025internvl3_5} assigns a score of 1 or 0 to each based on the provided reference image and video.
(2) Semantic Consistency ($\text{CLIP}_{re}$), the CLIP-score~\cite{radford2021learning} between the original text prompt and a VLM-generated caption of the generated video; (3) Hand Quality (HQ), assessed by the product of hand dynamics score and hand Laplacian sharpness score~\cite{lin2024cyberhost}; (4) Object Quality (OQ), assessed by the product of object dynamics score~\cite{zhang2025speakervid} and object DINO consistency~\cite{zhang2022dino}; and (5) Pixel-Level Interaction (PI), which verifies contact between DINO object boxes and DW-Pose keypoints. Our evaluation is further complemented by standard metrics. Ref Consistency with DINO for human appearance ($\text{DINO}_{subject}$) and ref image appearance ($\text{DINO}_{ref}$). 
Sync confidence ($\text{Sync}_{conf}$) for audio-visual consistency~\cite{Chung16a}, CLIP-B-T ($\text{CLIP}_{text}$) for text-video alignment, temporal consistency (Temp-C) and dynamic degree (DD) from VBench~\cite{zhang2024evaluationagent}.
Please refer to the supplementary materials for more benchmark details.
\begin{table*}
\vspace{-4mm}
\caption{\textbf{Quantitative comparison results on GroundedInter.} With different input signals, our single model can infer with various modes. TA2V means audio-driven, TAM2V means audio and motion-driven, and T2MV represents the use of self-generated motion. }
\vspace{-6mm}
\begin{center}
\scalebox{0.80}{
\footnotesize
\begin{tabular}{cccccccccccccc}
\toprule
\multirow{2}*{Task} & \multirow{2}*{Method} & \multicolumn{5}{c}{Human-Object Interaction} & \multicolumn{2}{c}{Ref Consistency} & \multicolumn{2}{c}{Text-Video Align} & \multicolumn{2}{c}{Audio and Video} \\
\cmidrule(r){3-7} \cmidrule(r){8-9} \cmidrule(r){10-11} \cmidrule(r){12-13}
& & VLM-QA $\uparrow$ & HQ $\uparrow$ & OQ $\uparrow$ & DD $\uparrow$ & PI $\uparrow$  & $\text{DINO}_{subject}$ $\uparrow$  & $\text{DINO}_{ref}$ $\uparrow$ & $\text{CLIP}_{text}$ $\uparrow$ & $\text{CLIP}_{re}$ $\uparrow$ & Temp-C $\uparrow$  & $\text{Sync}_{conf}$ $\uparrow$\\
\midrule
\multirow{4}*{Audio-Driven} & Wan-S2V~\cite{gao2025wan} & 24.65 & 0.336 & 0.063  & 0.095 & 0.619 & {0.670} & \textbf{0.870} & 0.281 & 0.885 & \textbf{0.955}  & 5.43\\ 
&Hunyuan-Avatar~\cite{chen2025hunyuanvideo} & \underline{26.88} & \underline{0.745} & {0.112} & \underline{0.321} & {0.666} & 0.661 & 0.841 & \underline{0.285} & 0.873& 0.934  & \underline{5.98}\\ 
&Fantasy-Talking~\cite{wang2025fantasytalking} & 25.35 & 0.416 & 0.098 & 0.127 & 0.593 & \textbf{0.673} & {0.855} & 0.276 & \underline{0.889} & {0.941}  & 5.79 \\ 
&OminiAvatar~\cite{gan2025omniavatar} & 26.76 & 0.732 & \underline{0.118} & 0.305 & \underline{0.732} & \textbf{0.639} & {0.847} & 0.282 & \underline{0.885} & {0.940}  & 5.95 \\ 
&Ours (TA2V) & \textbf{27.32} & \textbf{0.931} & \textbf{0.133} & \textbf{0.765} & \textbf{0.803} & \underline{0.671} & \underline{0.857} & \textbf{0.286} & \textbf{0.899} & \underline{0.945} & \textbf{6.04} \\ 
\midrule
\multirow{2}*{Pose-Driven} & UniAnimate-DiT~\cite{wang2025unianimate} & 24.65 & 0.862 & 0.082 & 0.441 & 0.669 & 0.621 & 0.782 & 0.277 & 0.875 & 0.938  & --\\ 
& Ours (TAM2V) & \textbf{28.47} & \textbf{0.957} & \textbf{0.141} & \textbf{0.773} & \textbf{0.832} & \textbf{0.648} & \textbf{0.842} & \textbf{0.287} & \textbf{0.902} & \textbf{0.940} & 5.73 \\ 
\midrule
\multirow{4}*{Text-Driven} & VACE~\cite{jiang2025vace} & 26.74 & 0.908 & 0.118 & 0.719 & 0.705 & 0.629 & 0.817 & 0.285 & 0.893 & \underline{0.940}  & -- \\ 
& HuMo~\cite{chen2025humo} & 24.12 & 0.910 & 0.101 & 0.380 & 0.491 & 0.566 & 0.726 & 0.281 & 0.880 & 0.935 & 5.15 \\ 
& Ours (T2MV) & \underline{29.05} & \textbf{0.975} & \textbf{0.150} & \textbf{0.792} & \textbf{0.852} & \underline{0.637} & \underline{0.835} & \underline{0.289} & \textbf{0.904} & 0.932 & -- \\ 
& Ours (TA2MV) & \textbf{29.07} & 
\underline{0.973} & \underline{0.147} & \underline{0.783}  & \underline{0.850} &\textbf{0.641}& \textbf{0.839} & \textbf{0.290} & \textbf{0.904} & \textbf{0.942}  & \textbf{5.92}\\ 
\bottomrule
\end{tabular}
}
\label{table:main_res}
\end{center}
\vspace{-2mm}
\end{table*}

\subsection{Comparison with the State-of-the-Art}
As there are currently no methods specifically designed for the task of GHOI, we compare our method with state-of-the-art (SOTA) approaches from several adjacent domains. For audio-driven methods, we select Hunyuan-Avatar~\cite{chen2025hunyuanvideo}, Wan-S2V~\cite{gao2025wan}, OmniAvatar~\cite{gan2025omniavatar}, and Fantasy-Talking~\cite{wang2025fantasytalking} for comparison. For subject-consistent generation, we compare against HuMo~\cite{chen2025humo}. For pose-driven animation, we benchmark against UniAnimate-DiT~\cite{wang2025unianimate}. VACE~\cite{jiang2025vace} is also utilized for comparison.
As shown in Tab. \ref{table:main_res}, the \textbf{audio-driven methods} demonstrate an advantage in ref consistency metrics but perform poorly on human-object interaction, particularly on Hand Quality (HQ) and Object Quality (OQ) scores. This is attributable to their primary focus on facial dynamics, which only allows for the generation of simple, low-dynamic actions and largely avoids interactive behaviors, also confirmed by their low dynamic degree (DD) scores. While this low dynamism facilitates the preservation of background and subject consistency, the absence of meaningful action is clearly reflected in the poor interaction scores. 
In contrast, our TA2V variant achieves a substantial lead in interaction metrics, representing an improvement of approximately 180\% in HQ, 111\% in OQ compared to the strong baseline Wan-S2V. Crucially, this is achieved while maintaining comparable reference preservation in terms of $\text{DINO}_{subject}$ and $\text{DINO}_{ref}$. On lip synchronization performance, our method even outperforms all audio-driven method in the HOI scenario.
For \textbf{pose-driven methods}, UniAnimate-DiT takes motion sequences generated by our PIM as input. Even with this high-quality guidance, our TAM2V variant outperforms it across all metrics, achieving a particularly notable margin on the OQ score. This discrepancy arises because pose-driven methods do not account for object morphology, leading to interaction misalignments and unnatural object deformations. Their inherent limitations also include the difficulty of obtaining and aligning pose sequences with the ref image, as well as the limited speaking ability.
Regarding the \textbf{subject-consistent method}, HuMo generates an entirely new video with subjects borrowed from the ref image, rather than continuing from the original scene, resulting in lower scores on reference consistency. Furthermore, our method significantly surpasses HuMo in terms of interaction quality. This is because interactions generated by subject-consistent models tend to be stereotypical and fail to produce smooth transitions from a non-interactive to an interactive state.
Notably, all our variants share a single model with different inference modes, highlighting its versatility and practical utility.

\subsection{Qualitative Evaluation}
\begin{figure*}[t]
\vspace{-3mm}
    \centering
\includegraphics[width=\textwidth]{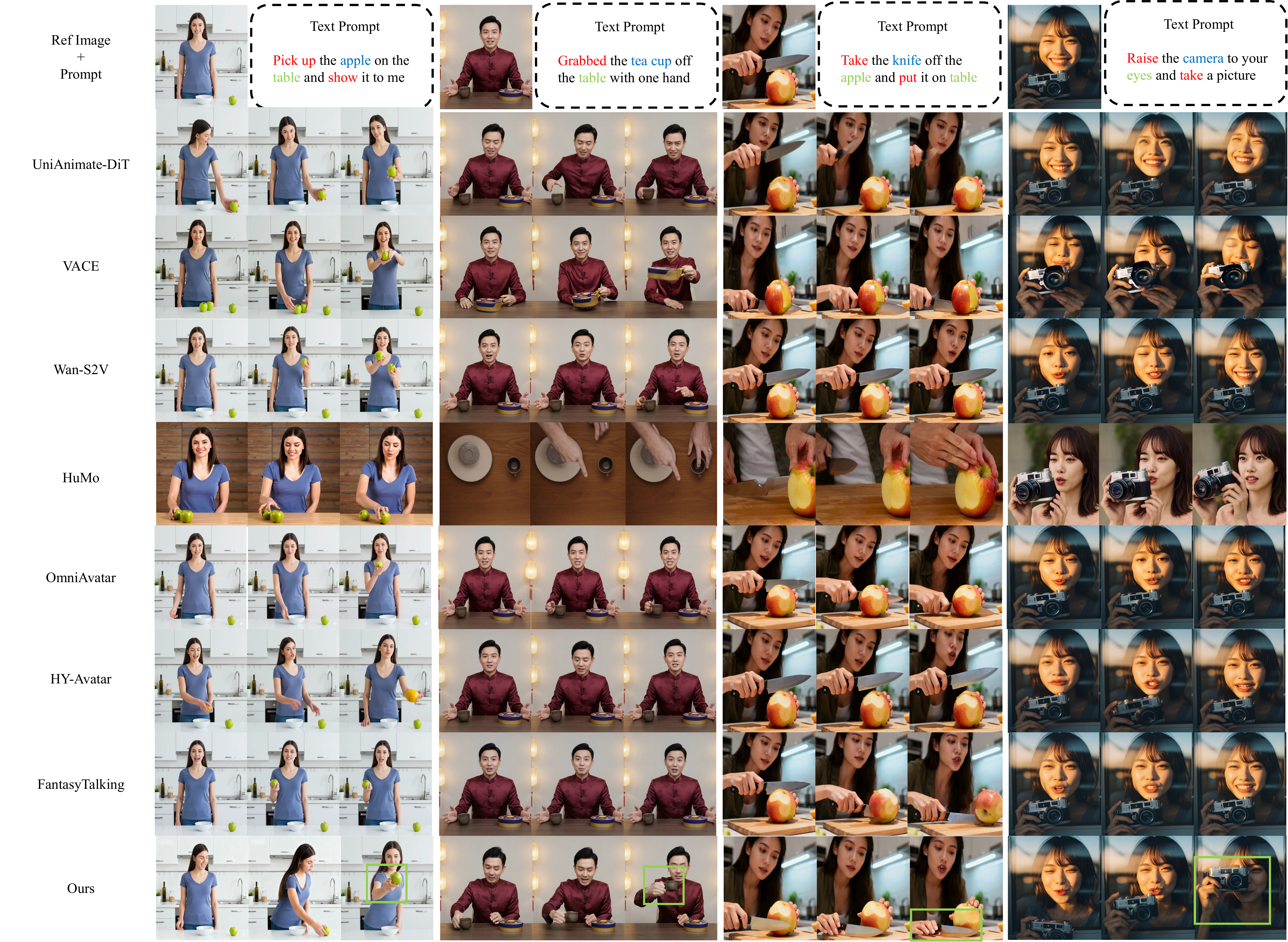}
\vspace{-6mm}
\caption{\label{fig:compare}\textbf{Qualitative comparison with SOTA methods.} Refer to Supp. Mat. for results of Hunyuan-Avatar and Fantasy-Talking.
\vspace{-4mm}
}
\end{figure*}
In Fig.~\ref{fig:compare}, we present a qualitative comparison between our method and SOTA approaches from related domains for the Grounded Human-Object Interaction task.
VACE exhibits responsiveness to text but suffers from severe artifacts during interaction, failing to preserve object integrity or accurately execute the interaction instructions. Moreover, it lacks the crucial audio-driven capability.
Wan-S2V successfully preserves the interaction environment specified by the reference image. However, its dynamism is largely confined to the audio-driven facial region, limiting its ability to respond to broader interaction instructions.
HuMo excels at generating fine-grained interaction details and facial dynamics. However, it fails to maintain the original interaction environment, instead using the subject from the reference image as an appearance template to synthesize an entirely new video, which is unsuitable for the GHOI task.
In contrast, our method accurately follows textual instructions, enabling the human to interact stably and plausibly with objects within the reference image environment. This is facilitated by the explicit decoupling of perception and planning from video synthesis. The PIM plans plausible scene-aware interaction motions, while the AIM is responsible for synthesizing the photorealistic details of the interactions.

\subsection{Ablation Studies}
\begin{table}
\caption{\textbf{Effect of Environmental Perception Training.} 
}
\vspace{-6mm}
\begin{center}
\scalebox{0.77}{
\footnotesize
\begin{tabular}{ccccccccc}
\toprule
Rope & Det. & Det.+Cont. & Last & VLM-QA $\uparrow$  & $\text{CLIP}_{re}$ $\uparrow$ & HQ $\uparrow$ & OQ $\uparrow$ & PI $\uparrow$\\
\midrule
 & & &   & 26.36 & 0.895 & 0.826 & 0.118 &0.711 \\ 
 & & & \checkmark  & 26.71 & 0.897 & 0.818 &0.115 & 0.710 \\ 
\checkmark &  &   &    & 27.15 & 0.902 & 0.845 & 0.123 & 0.730\\ 
\checkmark & \checkmark & &  & 27.87 & 0.902 & 0.917 & 0.132 & 0.751 \\ 
\checkmark & \checkmark & \checkmark& &  \textbf{28.89} & \textbf{0.904} & \textbf{0.925} & \textbf{0.144} & \textbf{0.780} \\ 
\bottomrule
\end{tabular}
}
\label{table:perception}
\vspace{-8mm}
\end{center}
\end{table}

\begin{table*}
\begin{center}
\vspace{-3mm}
\caption{\textbf{Ablation of PIM (top) and the Multimodal Training Strategy (bottom).} }
\vspace{-3mm}
\scalebox{0.85}{
\footnotesize
\begin{tabular}{cccccccccccccc}
\toprule
Method  & VLM-QA $\uparrow$ & HQ $\uparrow$ & OQ $\uparrow$ & DD $\uparrow$ & PI $\uparrow$  & $\text{DINO}_{subject}$ $\uparrow$  & $\text{DINO}_{ref}$ $\uparrow$ & $\text{CLIP}_{text}$ $\uparrow$ & $\text{CLIP}_{re}$ $\uparrow$ & Temp-C $\uparrow$  & $\text{Sync}_{conf}$ $\uparrow$\\
\midrule
w/o PIM & 26.19 & 0.711 & 0.104  & 0.587 & 0.685 & {0.638} & {0.840} & 0.285 & 0.892 & {0.936}  & 5.41\\ 
Cascade & 28.19 & 0.876 & 0.137 & 0.730 & 0.746 & 0.642 & 0.841 & 0.289 & 0.901 & 0.940  & {5.27}\\ 
Last-layer& 28.31 & 0.880 & 0.141 & 0.752 & 0.769 & {0.643} & \textbf{0.849} & 0.289 & 0.901 & 0.938  & 5.32 \\ 
Addition & 28.46 & 0.903 & 0.134 & 0.751 & 0.777 & 0.630 & 0.846 & {0.286} & 0.887 & 0.935  & 5.14 \\ 
Ours & \textbf{28.89} & \textbf{0.925} & \textbf{0.144} & \textbf{0.761}  & \textbf{0.780} &\textbf{0.646}& {0.846} & \textbf{0.290} & \textbf{0.904} & \textbf{0.942}  & \textbf{5.43}\\ 
\midrule
$AM$ & 28.23 & 0.912 & 0.133 & 0.745 & 0.770 & 0.639 & 0.838 & 0.289 & 0.900 & 0.933 & 4.23\\ 
$M \rightarrow AM$ & 28.84 & 0.920 & 0.141 & \textbf{0.764} & 0.778 & 0.636 & 0.837 & \textbf{0.290} & 0.903 & 0.934  & 3.98 \\ 
$A \rightarrow AM$ & 28.02& 0.906 & 0.134 & 0.738 & 0.755 & 0.640 & 0.842 & 0.287 & 0.897 & 0.940  & \textbf{5.49} \\ 
$AI \rightarrow IAM$ (Ours) & \textbf{28.89} & \textbf{0.925} & \textbf{0.144} & {0.761}  & \textbf{0.780} &\textbf{0.646}& \textbf{0.846} & \textbf{0.290} & \textbf{0.904} & \textbf{0.942}  & {5.43}\\ 
\bottomrule
\end{tabular}
}
\label{table:epm}
\vspace{-4mm}
\end{center}
\end{table*}

\begin{figure*}[t]
    \centering
\vspace{-2mm}
\includegraphics[width=\textwidth, height=0.36\textwidth]{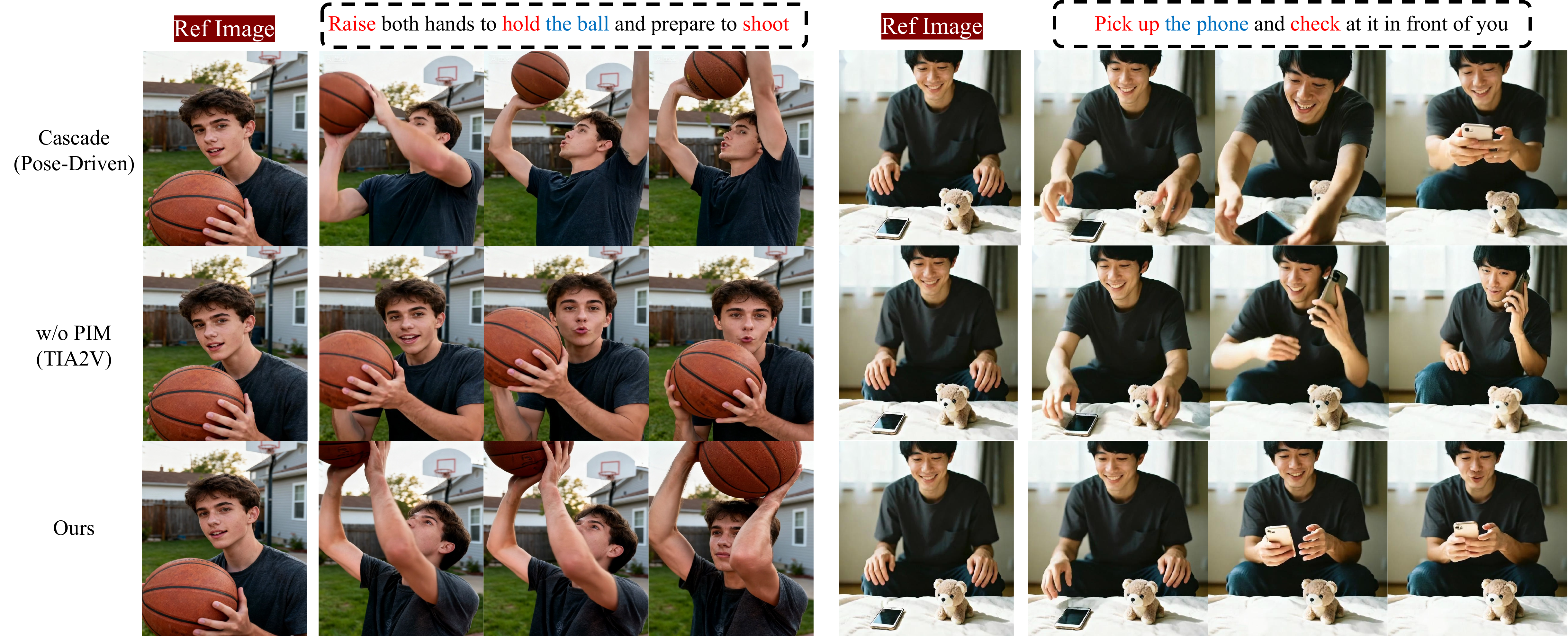}
\vspace{-6mm}
\caption{\label{fig:ablation}\textbf{Visualization of the PIM effect.} 
    }
\vspace{-6mm}
\end{figure*}

\textbf{Effect of Environmental Perception Training.}
We conduct an ablation study to evaluate different configurations of our environmental perception training process, with the results summarized in Tab.~\ref{table:perception}.
By default, the ref image is prepended to the video sequence. \textbf{Last} indicates that the ref image is appended after the last frame. \textbf{RoPE} specifies whether the modified Rotary Positional Embeddings (RoPE) are applied to the ref image. \textbf{Det.} denotes the use of detection data during PIM pre-training, which involves generating a single motion frame based on the ref image. \textbf{Det.+Cont.} represents generating a full motion sequence without ref motion.
Our findings indicate that prepending the ref image yields superior performance compared to appending it, due to the structural similarity between the reference image and motion frames. However, this advantage is contingent upon modifying RoPE to mitigate the excessive influence of the ref image on the generated frames.
Furthermore, we observe that detection data is crucial for establishing environmental perception abilities, while Det.+Cont. data serves as a vital bridge between perception and generation. 
Both types of data benefit the training of PIM.

\textbf{Effect of PIM.}
The homologous architecture of our PIM and AIM establishes intrinsic alignment in their feature representations, which we leverage for effective information fusion. As detailed in Tab.~\ref{table:epm} (top), we conduct an ablation study on the injection of interaction motion features. w/o PIM refers to training a single AIM as a text-image-audio-driven video generator, which exhibits relatively poor text alignment and human-object interaction quality. This results in the absence of environmental perception capabilities and exacerbates the control-quality dilemma in GHOI.
With PIM, we observe that a unified joint generation significantly outperforms a cascaded pipeline, in which a separately trained (frozen) PIM generates a motion sequence to drive the AIM. The cascaded approach creates an information bottleneck. Its reliance on explicit motion conditions is too restrictive to capture object shape changes, which in turn leads to artifacts and reduces the plausibility of the generated hands. In contrast, joint training allows for a richer flow of information.
Within the joint training paradigm, injecting only the last-layer feature from PIM into all AIM block layers mirrors the performance of the cascaded model, suggesting that globally applied motion is less effective than adaptive, layer-wise guidance. Our layer-wise residual injection mechanism, augmented with a zero-initialized linear layer, also outperforms simple layer-wise addition, confirming the advantages of our design.

As shown in Fig.~\ref{fig:ablation}, we visualize the effects of our PIM. With frozen PIM, the cascaded variant can be regarded as a pose-driven model, while w/o PIM corresponds to a TIA2V model. The pose-driven variant can generate text-aligned interaction videos but is prone to abnormal object deformations. The TIA2V variant often struggles with semantic comprehension, resulting in inaccurate execution of requested interactions, whereas our method successfully achieves reasonable and text-aligned interaction videos.

\begin{table}
\begin{center}
\caption{\textbf{Motion Representation Ablation}.}
\vspace{-4mm}
\scalebox{0.7}{
\footnotesize
\begin{tabular}{ccccccccccccc}
\toprule
 Method  & VLM-QA $\uparrow$ & HQ $\uparrow$ & OQ $\uparrow$ & DD $\uparrow$ & PI $\uparrow$  & $\text{DINO}_{sub}$ $\uparrow$ & $\text{Sync}_{conf}$ $\uparrow$\\
\midrule
2D (x,y) Rep. & 26.07 & {0.705} & {0.094} & 0.698 & 0.655 & 0.623 & {4.72}\\ 
2D (x,y) Cascade & {25.58} & 0.884 & 0.121 & 0.659 & {0.672} & {0.625} & 5.04 \\ 
Ours & \textbf{28.89} & \textbf{0.925} & \textbf{0.144} & \textbf{0.761} & \textbf{0.780} &\textbf{0.646} & \textbf{5.43}\\ 
\bottomrule
\end{tabular}
}
\label{table:network}
\vspace{-8.2mm}
\end{center}
\end{table}

\begin{table}
\begin{center}
\caption{\textbf{Real Scene Comparision.} }
\vspace{-4mm}
\scalebox{0.7}{
\footnotesize
\begin{tabular}{ccccccccccccc}
\toprule
Method  & VLM-QA $\uparrow$ & HQ $\uparrow$ & OQ $\uparrow$ & PI $\uparrow$   & $\text{DINO}_{sub}$ $\uparrow$ & $\text{CLIP}_{text}$ $\uparrow$ & $\text{Sync}_{conf}$ $\uparrow$\\
\midrule
Wan-S2V & 24.97 & 0.851 & 0.115 & 0.740 & {0.646} & 0.285 & 5.36\\ 
HY-Avatar & 25.29 & 0.817 & 0.108 & 0.724 & 0.644 & 0.285 & {5.45}\\ 
HuMo & 26.23 & 0.840 & 0.116 & 0.781 & {0.635} & 0.288 & 5.33 \\  Ours & \textbf{28.49} & \textbf{0.910} & \textbf{0.135} & \textbf{0.794} &\textbf{0.665} & \textbf{0.289} & \textbf{5.51}\\ 
\bottomrule
\end{tabular}
}
\label{table:real}
\vspace{-7mm}
\end{center}
\end{table}

\textbf{Motion Representation.} In Tab.~\ref{table:network} (top), we compare our RGB motion representation with a simpler 2d coordinate representation. 
The coordinate motion generation model lacks
priors and detection ability (generating motion requires an initial box and joint location)
, and generalizes poorly. 
Its huge modality gap with video hinders alignment (most driven methods use RGB), while cascaded ``generate (x,y) then render RGB for driven'' markedly underperforms joint generation.
RGB motion leverages video generation priors, improving text-to-motion accuracy and generalization without requiring large-scale datasets.
Shared RGB space simplifies video-motion alignment and reduces training complexity, yielding better results for GHOI generation.


\textbf{Real Scene Evaluation.}
For better SOTA comparison, we collect 50 cases from real scenes to build an additional test set. Quantitative experimental results on Tab.~\ref{table:real} demonstrate that our method performs well in real scenes.

\textbf{Multimodal Training Strategy.}
This section investigates the optimal strategy for sequencing multimodal conditions during AIM training. As detailed in Tab.~\ref{table:epm} (bottom), $A \rightarrow AM$ denotes first training on audio, followed by joint training on audio and motion. $AM$ represents training both modalities from the beginning, without any AIM audio pre-training.
The $A \rightarrow AM$ strategy achieves substantially better performance on audio metrics than either joint training from the beginning ($AM$) or pre-training on the motion modality first ($M \rightarrow AM$). This outcome stems from the asymmetric nature of the conditioning signals.
Consistent with the conclusions of Omnihuman~\cite{lin2025omnihuman}, audio is a comparatively weaker signal than interaction motion and requires a dedicated pre-training stage for effective learning; otherwise, its influence is overshadowed by the dominant motion signal.
Additionally, we find that enriching the audio-driven pre-training stage with Text-Image-to-Video data (without paired audio) yields notable benefits. This ensures that the model retains general motion dynamic priors, providing a stronger foundation for learning complex interaction motion during the final joint training phase.

\begin{figure}[h]
    \vspace{-4mm}
    \centering
\includegraphics[width=0.8\columnwidth,height=3cm]{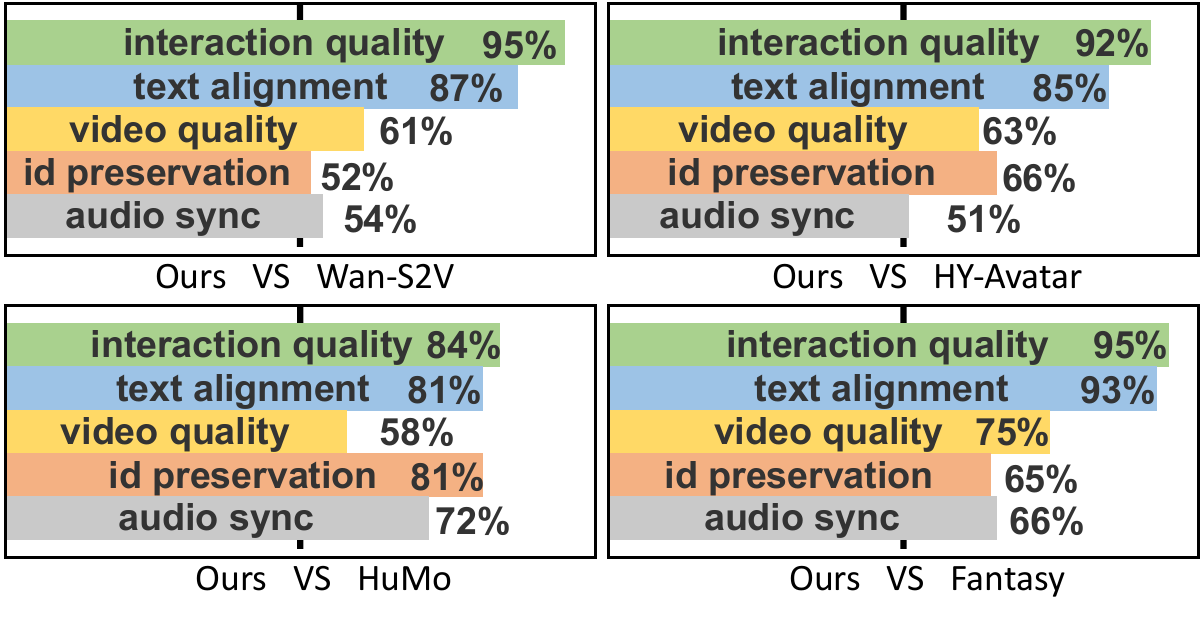}
     \vspace{-4mm}
\caption{\label{fig:user}\textbf{User Study Results.}}
\vspace{-6.6mm}
\end{figure}

\textbf{User Study}. We conduct a user study with 20 participants in 50 generation samples (each method). Given two videos generated from different methods, participants are asked to choose the better one. Results in Fig.~\ref{fig:user}.


\vspace{-2mm}
\section{Conclusion}
\vspace{-1mm}
This paper introduces a dual-stream talking avatar video generation framework that enables the generation of grounded human-object interactions.
Our method decouples perception and planning from video synthesis, enabling the generation of plausible, controllable, and high-quality GHOI videos.
The PIM is carefully designed to perceive the context of ref image and plan text-aligned interaction motion, while AIM generates vivid talking and interacting avatars with an M2V aligner. A benchmark is proposed for evaluating GHOI generation.
Comprehensive experiments showcase our model’s performance and highlight each module’s contributions and impacts. Our work provides valuable insights for future research in this field. 
The main limitation of our work is that it can only handle a single-person scene and is unable to generate multi-person involved HOI.

{
    \small
    \bibliographystyle{ieeenat_fullname}
    \bibliography{main}
}
\end{document}